\documentclass[conference]{IEEEtran}
\IEEEoverridecommandlockouts

\makeatletter
\def\ps@IEEEtitlepagestyle{%
  \def\@oddfoot{\mycopyrightnotice}%
  \def\@evenfoot{}%
}
\def\mycopyrightnotice{%
  {\footnotesize 979-8-3195-2824-7/27/\$31.00~\copyright~2027 IEEE\hfill}%
  \gdef\mycopyrightnotice{}
}
\makeatother

\usepackage{eso-pic}
\newcommand\AtPageUpperMyright[1]{\AtPageUpperLeft{%
 \put(\LenToUnit{0.17\paperwidth},\LenToUnit{-2cm}){%
     \parbox{0.9\textwidth}{\raggedleft\fontsize{8}{11}\selectfont #1}}%
 }}%
\newcommand{\conf}[1]{\AddToShipoutPictureBG*{\AtPageUpperMyright{#1}}}

\usepackage{cite}
\usepackage{amsmath,amssymb,amsfonts}
\usepackage{graphicx}
\usepackage{textcomp}
\usepackage{xcolor}
\usepackage{xurl}
\usepackage{booktabs}
\usepackage{array}
\usepackage{tabularx}
\usepackage{makecell}
\usepackage[hidelinks]{hyperref} 
\def\BibTeX{{\rm B\kern-.05em{\sc i\kern-.025em b}\kern-.08em
    T\kern-.1667em\lower.7ex\hbox{E}\kern-.125emX}}
\begin{document}

\title{\vspace*{1cm}
    From Regulation to Implementation: A Critical Evaluation of LLM-Assisted Regulatory Compliance in Industry
}

\author{
\IEEEauthorblockN{Marco B\"ucheler}
\IEEEauthorblockA{\textit{School of Engineering Technology} \\
\textit{Purdue University}\\
West Lafayette, United States \\
mbuechel@purdue.edu}
\and
\IEEEauthorblockN{Adriana Watson}
\IEEEauthorblockA{\textit{School of Engineering Technology} \\
\textit{Purdue University}\\
West Lafayette, United States \\
watso213@purdue.edu}
\and
\IEEEauthorblockN{Grant Richards}
\IEEEauthorblockA{\textit{School of Engineering Technology} \\
\textit{Purdue University}\\
West Lafayette, United States \\
grichard@purdue.edu}
}

\maketitle

 \conf{\textit{  Proc. of International Conference on Artificial Intelligence, Computer, Data Sciences and Applications (ACDSA 2027) \\
 2-4 February 2027, Rio de Janeiro-Brazil}}

\begin{abstract}
The European Union (EU) has emerged as a leading regulatory body in the development of sustainability and privacy regulations. While new regulatory requirements vary, many include a documentation artifact to ensure compliance. Notably, the Ecodesign for Sustainable Products Regulation (ESPR) introduces Digital Product Passports (DPPs) for life cycle transparency, while the General Data Protection Regulation (GDPR) mandates Data Protection Impact Assessments (DPIAs) to mitigate privacy risks. Creating these compliance artifacts, however, is challenging. Industrial data, which often exists in heterogeneous formats and is scattered across company and supplier systems, is required for DPPs and can be difficult to extract into compliant DPP formatting. Furthermore, DPIA documents require interdisciplinary expertise and follow no standardized format, making development difficult for novel systems. To address the particular complexity of compliance artifact creation for both regulations, researchers have proposed the use of LLMs in the generation process; however, the impact of the aforementioned problems on the output of these systems is largely unaddressed. This work investigates the existing research gap by exploring how data extraction instructions and regulatory vagueness impact the quality and consistency of LLM-produced compliance artifacts. The resulting artifacts are evaluated by benchmarking different models against manually created ground-truth schemas. The results reveal that less strict guidelines, such as DPIA formatting, require higher-context prompts to maintain consistency and completeness. For the more structured Digital Battery Passport (DBP) case, regulatory completeness remained comparatively stable, and 63.2\% of generated outputs conformed to the AAS 3.2 JSON schema, although additional structural validation identified remaining implementation errors.
\end{abstract}


\section{Introduction}
\label{sec:intro}
In recent years, the European Union (EU) has led the way in the development and implementation of both environmental and technological regulations. Notably, the Ecodesign for Sustainable Products Regulation (ESPR) and General Data Protection Regulation (GDPR) have distinguished themselves as leading regulatory works that address data rights and product sustainability, respectively \cite{european_commission_regulation_2024, european_commission_general_2018}. 

As a by-product of the novelty of these regulations, researchers have voiced growing concerns regarding the industrial implementation of these regulations \cite{politou_forgetting_2018, hauselmann_right_2024, king_proposed_2023}. Notably, a key point of conflict can be found in the challenges associated with producing the required documentation to conform to regulations; we henceforth refer to these documents as compliance artifacts. Although compliance artifacts such as Digital Product Passports (DPP) for the ESPR and Data Protection Impact Assessments (DPIAs) for the GDPR are essential for both compliance and accountability, they each pose unique challenges that make them difficult to produce manually. The proliferation of Large Language Models (LLMs) has brought forth the opportunity to develop novel solutions for the automation of key compliance artifacts. However, current research does not properly address how differences in prompting approaches impact the effectiveness of compliance artifact generation. We address this gap by first identifying the novel ways in which LLMs are being used to produce compliance artifacts and second, reproducing these methods with varying levels of prompt specificity in a case study to identify the extent to which LLM-generated compliance artifacts vary given differences in prompting. Notably, we select two distinct compliance artifacts to examine in the case study from two ends of the compliance spectrum: one clearly defined but not yet deployed standard and one vaguely defined but fully deployed standard. This serves to represent the extreme ends of existing regulatory requirements as they relate to artifact creation mandates.

\section{Background and Related Work} \label{sec:background}

\subsection{Compliance Artifacts} \label{sec:regs}
\subsubsection{ESPR/EU Battery Regulation}
In the EU, the concept of DPPs was first introduced by the European Green Deal \cite{european_commission_european_2019} and the Circular Economy Action Plan \cite{european_commission_new_2020}. These passports are compliance artifacts designed to enable transparency and circularity for products sold in the European market. The ESPR regulation defines requirements for DPPs by outlining the structure and content needed in each artifact. To be considered a valid artifact, DPPs must be structured as a digital product record built on open standards. The passports must be directly linked to a physical product via a data carrier, such as a barcode or QR code. The carrier must link to a persistent, unique product identifier to ensure end-to-end life cycle traceability. While the exact data fields will be determined by product-specific delegated acts, the ESPR establishes a baseline for artifact compliance. 


The EU Battery Regulation introduces Digital Battery Passports (DBPs) with the primary objective of minimizing the environmental impact of batteries \cite{european_commission_regulation_2023, battery_pass_consortium_battery_2023}. As a result, DBPs require comprehensive data on a battery's identity, material composition (including recycled content and hazardous substances), and carbon footprint. Moreover, performance and durability metrics, detailed instructions for repair, repurposing guidelines, and end-of-life recycling need to be documented \cite{european_commission_regulation_2023, battery_pass_consortium_battery_2023}.
\subsubsection{GDPR}
While the ESPR is concerned with the life cycle of physical products, the General Data Protection Regulation (GDPR) is a comprehensive rulebook on data life cycles. Notably, it defines the requirements of data controllers for the collection, use, modification, storage, and deletion of data. 


The GDPR, like the ESPR, also requires explicit documentation as a key compliance stage. Notably, the GDPR introduced the requirement for a DPIA in Article 35, intended to ensure data controllers dealing with high-risk data have done their due diligence to evaluate the data life cycle \cite{european_commission_general_2018}. Specifically, GDPR Article 35(7) outlines the following documentation expectations for the DPIA: a description of the reasons for collecting the data and how it will be used, an analysis of the appropriateness of the data processing as it relates to the previously cited purpose, an analysis of the risks associated with the data processing operations, specifically as they relate to the data subject's data rights, a thorough analysis of the risk mitigation protocols that will be used to maintain data subjects' security, privacy, and data rights.  

\subsection{AI Systems for Regulatory Compliance}
\label{sec:ai-compliance}

\subsubsection{DPP/DBP generation}
In industrial environments, data required for regulatory compliance is frequently scattered across company and supplier systems. Additionally, various data formats, including information technology (IT) and operational technology (OT) data, are needed. Building an EU-compliant product life cycle record (such as a DPP or DBP) requires data from various sources and heterogeneous formats. As a result, transforming and bundling existing data into a standardized format across supply chain partners is a necessary step in passport development. One standard that facilitates the generation of structured product life cycle representations is the Asset Administration Shell (AAS), which a variety of sources cite as an ideal candidate to create standardized passports \cite{pourjafarian_multi-stakeholder_2023, palm_architectures_2024}. The AAS offers interoperable data exchange possibilities, making it a promising solution for companies to comply with the EU's ESPR and Battery regulations. Data in the AAS is structured in submodels \cite{kuenster_opportunities_2023} which describe informational content or functional elements of a product (e.g., digital nameplate or carbon footprint submodel). Properties then define the characteristics of each submodel in detail.

Creating and populating administration shells is, however, a challenging and labor-intensive process for many companies. As a result, existing literature proposes the use of LLMs to automate AAS creation \cite{xia_generation_2024}. In such pipelines, LLMs extract core data attributes from industrial documents, perform semantic searches against established external dictionaries, and synthesize the information into structured JSON or XML formats compliant with AAS specifications. Empirical evaluations show that LLM-driven architectures can successfully automate the creation of error-free AAS instances with an effective generation rate of 62-79\% \cite{xia_generation_2024}.

\subsubsection{DPIA Generation}

Industrial systems are also heavily impacted by the GDPR, particularly in how they collect, use, and maintain data. While many solutions for GDPR compliance in data systems relate to more traditional database automation approaches, LLMs have gained popularity in the solution space for a variety of applications\cite{maguire_metadata-based_2015}, \cite{hublet_enforcing_2024}. 

The majority of existing literature focuses primarily on two key aspects of GDPR compliance: regulation interpretation and verification. 


Language models have also been proposed as a key tool in generating initial data collection artifacts, notably the DPIA. As the GDPR's specifications for the DPIA are relatively vague, many researchers have pointed to a great amount of variety in the interpretation and execution when producing this artifact \cite{dalla_corte_data_2022}. Although some authorities have developed software to aid in the development of DPIAs, these solutions are not universal across the EU nor required, as the GDPR allows data controllers to use a method of their choice \cite{korff_gdpr_2020}. Korff also notes a key tension in the creation of such a document; notably, companies are used to framing risk assessments around risks associated with the company: the DPIA is a risk assessment framed around the risk to individuals. Despite this tension, very little literature explores the use of automation techniques or LLMs to assist in the generation of these artifacts \cite{iacono_tool_2024}. In contrast, the use of LLMs to generate or assist in the creation of risk assessments more broadly is widely explored \cite{valkama_using_2025}, \cite{collier_how_2025}. Due to the complexity of DPIA generation, it seems worthwhile, then, to explore how LLMs may be used to enhance the uniformity and thoroughness of these artifacts. 

\subsection{Key Problems with LLM-Generated Artifacts}
\label{sec:trustworthiness}
Despite the promising nature of LLM-enhanced artifact generation, similar applications have encountered a variety of concerns that warrant caution. Famously, LLMs are known to hallucinate, or invent material rather than pulling from real sources \cite{ji_towards_2023}. In the context of AI-generated artifacts, this may result in the generation of invalid structures, information, or scenarios. 

Beyond hallucination, the creation of artifacts such as DPPs and DPIAs often requires specified domain knowledge. While LLMs may perform well in some scenarios due to their wide breadth of training data, this advantage may also cause interpretation problems, as some researchers have noted that LLMs struggle when it comes to interpreting or using specialized terminology or logic \cite{collier_how_2025}. 


All of these problems, however, do have mitigation strategies, including prompting strategies, combined LLM and RAG approaches, and the use of isolated models or federated learning \cite{ji_towards_2023, abualhaija_llm-assisted_2025, li_review_2020}. Thus, it is worth investigating how these artifacts can be generated more effectively using LLMs. 

\section{Case-Study Methodology}\label{sec:methodology}

To observe how LLMs handle different prompting approaches in the creation of compliance artifacts, we present a case study focused on the automated generation of an ESPR/EU Battery Regulation-compliant DPP and a GDPR-compliant DPIA. 

\subsection{Experimental Setup}\label{sec:setup}
Five distinct models, noted in Table~\ref{tab:models}, are selected to represent a variety of developers and accessibility. 

\begin{table}[!ht]
\centering
\caption{Overview of Evaluated Language Models}
\begin{tabular}{lcc}
\toprule
\textbf{Model} &  \textbf{Developer} & \textbf{Access}  \\
\midrule
GPT-4o                         & OpenAI        & API   \\
Claude Sonnet 4.6               & Anthropic     & API    \\
Meta-Llama-3.1-8B-Instruct       & Meta          & Open  \\
Ministral-8B-2512       & Mistral AI    & Open         \\
Qwen2.5-7B-Instruct               & Alibaba   & Open        \\
\bottomrule
\end{tabular}%
\label{tab:models}
\end{table}

The models are all prompted with a temperature of 0.7 and a 32k token cap, at which point model outputs were truncated. Each run used the same structured input, described in Section~\ref{sec:prompting-strategy}. Each model runs 20 times for each task and vagueness level. As a result, 800 total outputs are collected: 2 tasks (DPP and DPIA) $\times$ 4 prompt vagueness levels (baseline, high, medium, low) $\times$ 5 models $\times$ 20 runs.

\subsection{Prompting Strategy}
\label{sec:prompting-strategy}
The models are prompted with a structured input based on the input task and predefined vagueness levels. Table~\ref{tab:prompt-levels} describes the content added at each vagueness level for both scenarios.

\begin{table}[b!]
\centering
\caption{Context added at each vagueness level beyond the shared core
(role, product data or processing scenario, JSON-only output).
Levels are not cumulative.}
\label{tab:prompt-levels}
\footnotesize
\setlength{\tabcolsep}{3pt}
\renewcommand{\arraystretch}{1.15}
\begin{tabularx}{\columnwidth}{@{}l
  >{\raggedright\arraybackslash}X
  >{\raggedright\arraybackslash}X@{}}
\toprule
\textbf{Level} & \textbf{DBP prompt adds} & \textbf{DPIA prompt adds} \\
\midrule
Low      & Names the AAS Material Composition submodel; model decides
           which fields are required
         & Model decides which sections are required \\
Baseline & Battery Regulation text
         & Art.~35(7)(a)--(d) and Recital~84 verbatim; one key per
           sub-paragraph \\
Medium   & Target-product disambiguation; extraction rules (preserve
           structure, metric units, no invented values); list of
           mandatory data
         & Plain-language summary of the four Art.~35(7) components \\
High     & Annotated regulatory text; full IDTA 02035-6 template;
           AAS/DPP domain context; extraction rules
         & Four mandatory sections expanded into 21 required items
           (e.g., lawful basis, retention, likelihood/severity,
           residual risk) \\
\bottomrule
\multicolumn{3}{@{}p{\columnwidth}@{}}{\scriptsize Full prompts:
  \url{https://github.com/awatson246/evaluating-llm-generated-compliance-artifacts.git/}} \\
\end{tabularx}
\end{table}

\subsection{Evaluation}
\label{sec:evaluation}
To evaluate the outputs of the models, a gold standard schema is created for the DBP and DPIA scenarios, respectively\footnote{While the gold standard schema is not included here due to length restrictions, it is available in the linked codebase for the experiment.}. These schemas are based on structures and examples present in existing legal, industrial, and academic documentation for each regulation to reflect current best practices. 

\subsubsection{Schemas}
The DBP schema is publicly available through the Industrial Digital Twin Association (IDTA) \cite{industrial_digital_twin_association_ev_digital_2026}. The name of the submodel used for this experiment is ``Digital Battery Passport \- Part 6: Material Composition (IDTA-02035-6)''. The values of an exemplary industrial battery is used and placed into the submodel template as the reference file to which the LLM outputs are compared. 

The DPIA schema combines two published templates, one with a more narrative structure and one with a more linear structure \cite{dalla_corte_data_2022} \cite{noauthor_tools_2024}. It includes key questions from both templates in several sections with both Boolean and written response outputs. The full schema is available in the linked codebase. Each field includes a ``required'' boolean marker that indicates whether the information is required explicitly by the GDPR, or simply recommended for best practices based on existing templates and documentation. 

\subsubsection{Methodology for Analysis}
The outputs are then evaluated on the basis of two key factors: consistency and compliance with regulation. Consistency is determined by assessing the structure and content of each run of the same type (same task, model, and vagueness level) to determine how stable the model outputs are given the same setup. A field stability score is calculated for each field present in the output of each run to determine the presence of a given field across multiple runs. A field is considered stable if it is present in all runs (a stability score of 1.0).  

To evaluate compliance with the regulation, the gold standard schema is used to compare each consolidated task/vagueness/model output against regulatory requirements. The components correlating to each section of the appropriate schema are extracted from the model output and combined from outputs of the same task, vagueness level, and model. The information is then compared against the content required by the gold-standard schema, and a regulatory compliance score is calculated on the basis of the inclusion and completeness of required components. An assessment of the commonly excluded and included information for each model is then conducted for both regulatory tasks. 

For the ESPR/EU Battery Regulation task, a third evaluation is introduced: technical AAS validity, distinguishing the inclusion of required regulatory information from the technical correctness of the generated AAS artifact. Technical validity is evaluated at three stages. First, is the artifact a syntactically valid JSON; second, is it in conformance with the AAS 3.2 JSON schema; and third, a full validation pass that additionally evaluates structural and consistency requirements (of the AAS environment).

\section{Results}
\subsection{ESPR/EU Battery Regulation Compliance}\label{sec:espr-results}
\subsubsection{Consistency}
As indicated by Table \ref{tab:espr_consistency}, models performed only slightly differently across vagueness levels. Qwen-2.5 7B and GPT-4o maintained perfect cross-run consistency ($1.00$) at every context level. Llama 3.1 8B also achieved perfect cross-run consistency for all completed context levels. The perfect cross-run consistency observed for these models suggests that they resolve ambiguous inputs through a stable, deterministic extraction strategy. Mistral 8B also follows this trend but demonstrates a slight dip in performance at low context. Claude Sonnet 4.6, by contrast, is the only model to exhibit meaningful drops in consistency, falling to $0.96$ at low context and reaching a minimum of $0.77$ given medium context. The reduced consistency at medium context indicates that Claude Sonnet 4.6 produced a structurally different output in roughly one in four repeated runs for the same moderately vague prompt. Notably, Claude Sonnet 4.6 recovered to full consistency ($1.00$) when given the high-context prompt.

\begin{table}[!ht]
  \centering
  \caption{Cross-run consistency scores for DBP artifacts by model and vagueness level.}
  \begin{tabular}{lccccc}
    \toprule
    \textbf{Context} & \makecell{\textbf{Qwen-2.5}\\ \textbf{7B}} & \makecell{\textbf{Claude} \\\textbf{Sonnet 4.6}} & \makecell{\textbf{GPT-}\\\textbf{4o}} & \makecell{\textbf{Llama}\\ \textbf{3.1 8B}} & \makecell{\textbf{Mistral} \\\textbf{8B}} \\
    \midrule
    Baseline & 1.00 & 1.00 & 1.00 & 1.00 & 1.00 \\
    Low & 1.00 & 0.96 & 1.00 & 1.00 & 0.96 \\
    Medium & 1.00 & 0.77 & 1.00 & 1.00 & 1.00 \\
    High & 1.00 & 1.00 & 1.00 & -- & 1.00 \\
    \bottomrule
  \end{tabular}
  \label{tab:espr_consistency}
\end{table}

\subsubsection{Compliance}
At baseline, all five models achieved perfect compliance ($1.00$), confirming that fully specified inputs pose no challenge for any of the systems evaluated. Qwen-2.5 7B, GPT-4o, Mistral 8B, and Llama 3.1 8B maintained this perfect score across all context levels (with the exception of Mistral 8B, which dropped slightly at low context). 


When examining the required field inclusion rates by model, averaged across all vagueness levels, seven of the ten required fields are included at $100\%$ by nearly every model. Mistral 8B is the only exception, as it includes the \verb|Hazardous substance class|, \verb|Hazardous substance concentration|, and \verb|Hazardous substance identifier| 98\% of the time. 

\subsubsection{Technical Validity}

Regulatory completeness did not always translate into technically valid AAS artifacts. Of the 400 ESPR outputs, 329 (82.2\%) were valid JSON and 253 (63.2\%) conformed to the AAS 3.2 JSON schema. Among outputs that were valid JSON, 76.9\% were also AAS schema-valid. \newline Only 13 outputs (3.2\%) passed the complete validator, which includes additional AAS checks like element consistency, permitted properties, structural relationships, and idShort uniqueness.

\begin{table}[!ht]
\centering
\caption{Technical validity of generated ESPR artifacts. Percentages indicate the share of generated outputs passing each validation stage.}
\begin{tabular}{lccc}
\toprule
\textbf{Model} &
\makecell{\textbf{Valid}\\\textbf{JSON}} &
\makecell{\textbf{AAS 3.2}\\\textbf{Schema-Valid}} &
\makecell{\textbf{Full Validator}\\\textbf{Pass}} \\
\midrule
Qwen-2.5 7B           & 72.5\% & 66.2\% & 0.0\% \\
Claude Sonnet 4.6  & 75.0\% & 63.8\% & 11.2\% \\
GPT-4o             & 76.2\% & 58.8\% & 0.0\% \\
Llama 3.1 8B          & 97.5\% & 78.8\% & 5.0\% \\
Ministral-8B-2512  & 90.0\% & 48.8\% & 0.0\% \\
\midrule
\textbf{Overall}             & \textbf{82.2\%} & \textbf{63.2\%} & \textbf{3.2\%} \\
\bottomrule
\end{tabular}
\label{tab:aas-validity}
\end{table}

Llama 3.1 8B achieved the highest valid-JSON and AAS-schema validity rates, while Claude Sonnet 4.6 produced the most outputs passing the complete validator. AAS schema validity decreased from 98\% at baseline and 90\% at low context to 60\% at medium and 5\% at high context, indicating that additional prompt context did not consistently improve technical validity.

The most frequent validation issues were unexpected properties, schema violations, and element-consistency errors, while 71 outputs failed because of invalid JSON syntax.

\subsection{GDPR Compliance}\label{sec:gdpr-results}
\subsubsection{Consistency}
As indicated by Table~\ref{tab:gdpr_consistency}, models performed in vastly different ways given different context levels. Qwen-2.5 7B and Llama 3.1 8B returned equally consistent results across all context levels. Claude Sonnet 4.6, GPT-4o, and Mistral 8B demonstrated steady improvements in consistency, with a large improvement at the highest context level. 

\begin{table}[!ht]
  \centering
  \caption{Cross-run consistency scores for GDPR artifacts by model and vagueness level.}
  \begin{tabular}{lccccc}
    \toprule
    \textbf{Context} & \makecell{\textbf{Qwen-2.5}\\ \textbf{7B}} & \makecell{\textbf{Claude} \\\textbf{Sonnet 4.6}} & \makecell{\textbf{GPT-}\\\textbf{4o}} & \makecell{\textbf{Llama}\\ \textbf{3.1 8B}} & \makecell{\textbf{Mistral} \\\textbf{8B}} \\
    \midrule
    Baseline & 0.92 & 0.51 & 0.49 & 0.81 & 0.49 \\
    Low & 0.90 & 0.57 & 0.43 & 0.80 & 0.48 \\
    Medium & 0.91 & 0.58 & 0.57 & 0.83 & 0.62 \\
    High & 0.91 & 0.77 & 0.77 & 0.88 & 0.78 \\
    \bottomrule
  \end{tabular}
  \label{tab:gdpr_consistency}
\end{table}

\subsubsection{Compliance}
\begin{figure}[!ht]
    \centering
    \includegraphics[width=1.0\linewidth]{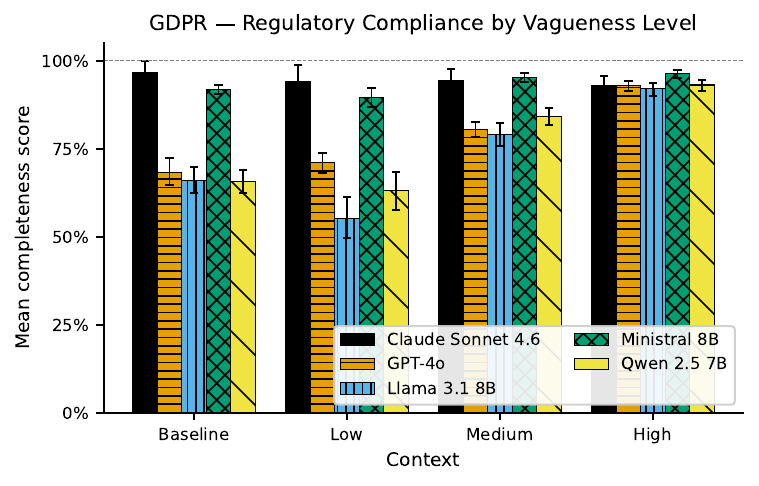}
    \caption{Mean regulatory compliance (completeness) scores for GDPR artifacts by model and vagueness level.}
    \label{fig:gdpr-completeness}
\end{figure}

Figure~\ref{fig:gdpr-completeness} demonstrates the mean completeness scores calculated based on the inclusion of required fields in the DPIA schema. For most models tested, the highest context prompts yielded the most compliant outputs with the smallest deviation. Claude Sonnet 4.6 and Mistral 8B produced stable, highly complete outputs across all context levels, while all other models returned more complete results with smaller error margins as more context was included in the prompt.

\begin{figure}[!ht]
    \centering
    \includegraphics[width=1.0\linewidth]{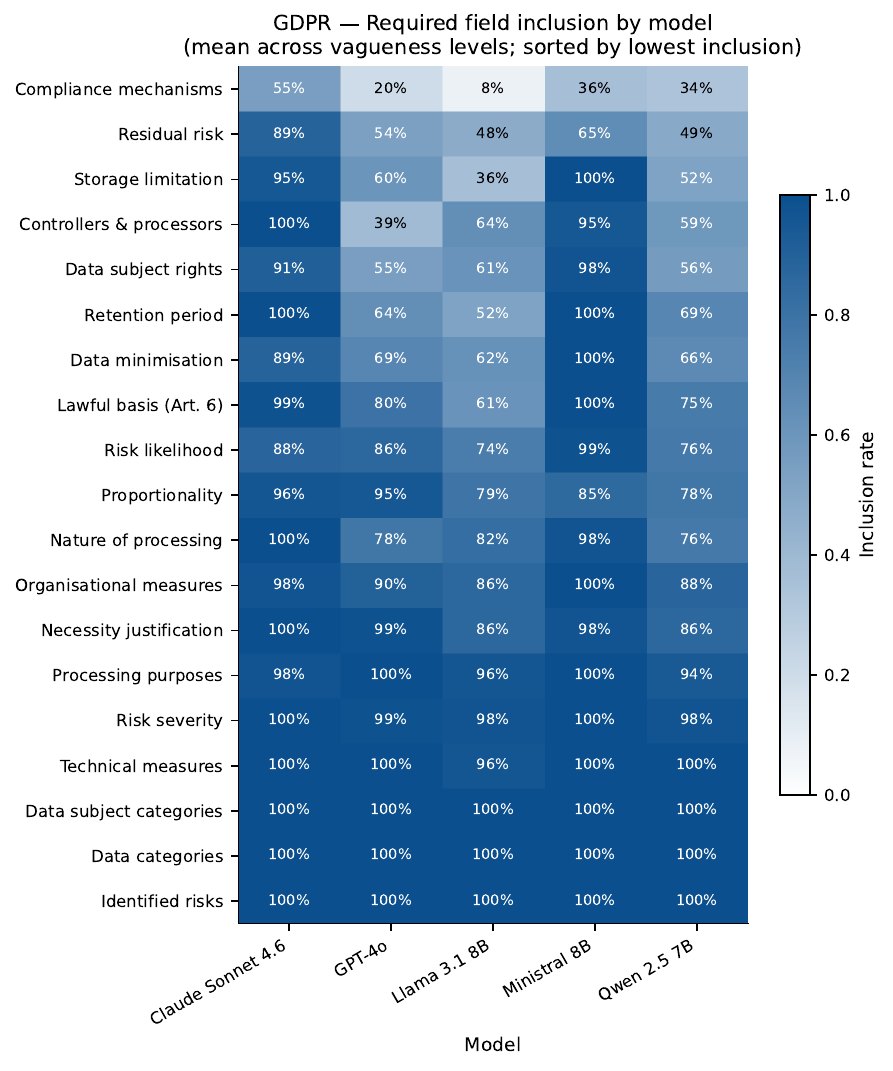}
    \caption{Required field inclusion rates (\%) for GDPR artifacts, averaged across all vagueness levels.}
    \label{fig:gdpr-compliance}
\end{figure}

Figure~\ref{fig:gdpr-compliance} demonstrates the inclusion rate of required fields separated by model and averaged across all context levels. While some fields, such as identified risks, data subject categories, data categories, and technical measures, are always included, others are frequently left out. Notably, compliance mechanisms are only included, on average, less than half the time, while residual risk, storage limitations, and controllers \& processors are included by some models and forgotten in others. Following earlier results, Claude Sonnet 4.6 and Mistral 8B included the most required fields. 

\section{Discussion}\label{sec:discussion}

\subsection{Implications for Regulation}\label{sec:implications}
The DBP results stand in notable contrast to those observed for the DPIA. Regulatory completeness remained high across most models and context levels. The results suggest that the concrete field structure of the AAS submodel reduces ambiguity in determining what information should be included. The finding is also reflected in the technical evaluation, which revealed that structured specifications tend to effectively guide LLMs toward the required content and format, but do not by themselves guarantee technically implementation-ready AAS artifacts. The decline in JSON and schema validity at higher context levels further indicates that providing more detailed instructions does not necessarily improve technical validity.

The DPIA results demonstrate a different challenge. Models generally produced more consistent outputs when additional contextual guidance was provided than when relying primarily on regulatory text. Regulatory completeness followed a similar pattern, with higher-context prompts generally capturing more of the information represented in the ground-truth schema. Frequently omitted components included compliance mechanisms, residual risk, data subject rights, storage limitations, and data minimization. These findings suggest that the requirements of GDPR Article 35 alone may not provide sufficient structure for consistent automated DPIA generation and that relevant requirements from the broader GDPR context may need to be incorporated into the generation process.

Together, the two cases indicate that the level of structure provided by a regulation or supporting specification influences how effectively LLMs can generate compliance artifacts. Granular specifications such as the DBP schema support comparatively consistent and complete outputs, although technical validation remains necessary. More principle-based requirements, such as those governing DPIAs, require additional contextual guidance to achieve comparable consistency and completeness. Future research should therefore investigate both how regulatory context can be structured for less-defined compliance artifacts and how LLMs can be guided toward generating technically valid AAS artifacts.

\subsection{Limitations}\label{sec:limitations}
The scope of the research, while detailed, only analyzes a small set of theoretical case studies that are designed based on existing scenarios. This is done to limit the scale of the study to verify the evaluation mechanisms and to avoid the use of private or otherwise protected data. Building out the case-study set to include real-world examples with in-use DBP and DPIA documentation would greatly increase the context and insights that could be gained. 

\section{Conclusion}\label{sec:conclusion}
As regulations become increasingly complex and ubiquitous, it is necessary to carefully examine the tools we use to comply with them. Although LLMs show great promise in closing the gap in regulatory complexity, the work presented in this paper clearly demonstrates that the nature of the prompt matters. While established frameworks, such as those available for DBPs, can circumvent the need for detailed prompts, practitioners must be more careful in defining instructions for less-defined documents such as the DPIA.

Although prompting strategy can have a significant impact on the consistency and completeness of outputs, more research is needed to properly define best practices for compliance artifact generation. Furthermore, to inspire more consistent compliance, the EU should consider formalizing output specifications in delegated acts and guidance documents.  

\section{Appendix}\label{app:code}
\subsection{Code}
All code, including prompts and outputs, is available here: \href{https://github.com/awatson246/evaluating-llm-generated-compliance-artifacts.git}{https://github.com/awatson246/evaluating-llm-generated-compliance-artifacts.git}

\bibliography{references.bib}

\end{document}